%% file: main.tex
\documentclass[runningheads]{llncs}

\usepackage[latin1]{inputenc}
\usepackage{francois-preamble}
\usepackage{cleveref}
\usepackage{subcaption}
\usepackage{blindtext}
\usepackage{wrapfig}
\usepackage{ifthen} 
\newboolean{Drafty}%
\setboolean{Drafty}{true}%
\ifthenelse{\boolean{Drafty}}%
{%
	\newcommand{\francois}[1]{{\color{red}\textbf{Francois: }#1}}
	\newcommand{\Sune}[1]{{\color{green}\textbf{Sune: }#1}}
	\newcommand{\kenny}[1]{{\color{brown}\textbf{Kenny: }#1}}
}%
{%
	\newcommand{\francois}[1]{{}}
	\newcommand{\Sune}[1]{{}} 
	\newcommand{\kenny}[1]{{}} 
	
}%

\title{First Order Locally Orderless Registration}
\author{}
\author{Sune Darkner\inst{1} \and
Jos{\'e} D.T. Vidarte \inst{1} \and
Fran\c{c}ois Lauze \inst{1}}

\authorrunning{Anonymous SSVM submission}
\institute{Department of Computer Science, University of Copenhagen\\\email{darkner,jota,francois@di.ku.dk}}

\begin{document}




\maketitle 

\input{abstract}
\input{introduction}
\input{previous_work}
\input{background}

\input{method}
\input{results}
\input{conclusion}

\input{acknowledgements}

%
%
\bibliographystyle{splncs04}
\bibliography{mybib}

\end{document}

%% file: abstract.tex
\begin{abstract}

First Order Locally Orderless Registration (FLOR) is a scale-space framework for image density estimation used for defining image similarity, mainly for Image Registration. The Locally Orderless Registration framework was designed in principle to use zeroth-order information, providing image density estimates over three scales: image scale, intensity scale, and integration scale. 
We extend it to take first-order information into account and hint at higher-order information. We show how standard similarity measures extend into the framework. We study especially Sum of Squared Differences (SSD) and Normalized Cross-Correlation (NCC) but present the theory of how Normalised Mutual Information (NMI) can be included.

\keywords{Image Registration  \and Locally Orderless Images \and First Order Information.}
\end{abstract}

%% file: introduction.tex
\section{Introduction}

Image similarity is generally based on zeroth-order information by a scalar to scalar comparison, e.g. Sum of Squared Differences (SSD), Normalised Cross-Correlation (NCC) or Mutual Information (MI)  \cite{klein2009evaluation}. However, images have structure and they encode information that extends beyond zeroth-order, they do not look like random noise. MI and NCC do incorporate more than just pixel intensity but very weakly and indirectly. Higher-order information is seldom used with a few exceptions, notably the normalized gradient fields \cite{haber2006intensity}. We aim to integrate high-order information for registration based on Locally Orderless Images (LOI) \cite{koenderink1999structure} and Locally Orderlles Registration LOR \cite{darkner2013locally}. 


LOI defines three fundamental scales for estimating a density from an image: the \emph{spatial}  scale, which is the "classical" scale-space one, the \emph{intensity or information} scale, as "bin scale" and the \emph{integration} scale, which define the localisation of the density estimates of intensity distributions. The key is to 'marginalize over the geometry' and leave only the correspondence of information. The locally orderless registration gives us a theoretical platform to perform this marginalization for scalar-valued images.


Locally Orderless Registration (LOR) \cite{jensen2015locally} explored its application for Magnetic Resonance Diffusion-Weighted Imaging (DWI), which are images containing complicated geometries. Indeed, DWI images can be seen as functions $\bI: \Omega\times\SS^2\to \RR$, with $\Omega$ an open subset of  $\RR^3$, where $\SS^2$ is seen as the space of directions (with orientation) in $\RR^3$.  An extra directional scale is added before building and localizing densities.


In this work we extend the LOI and LOR \cite{darknersporring2012pami,darknersporring2011ipmi,jensen2015locally} framework for images $\bI : \Omega\to \RR$, by lifting these images to images $\bI : \Omega\times\SS^2 \to \RR$, where $\SS^2$  this time parametrizes the local orientations of the image $\bI$. This is performed through directional responses of derivatives of Gaussian. Other kernels could be used, for instance, non-symmetric ones. Lifts to second or higher-order structures can similarly be defined via higher kernel derivatives. Once the lifting has been performed, ideas similar to DWI image registration can be used. However, as opposed to the DWI case, this lifting comes already with its scale parameter. This lifting idea is not new, with especially works in the context of image smoothing and disentangling of directions (\cite{janssen_etal:2018} and references therein). Tools and end goals in this work are different: classical Gaussian filters and image registration.

Given two images $I$ and $J$, the registration problem is to find the transformation $\phi: \RR^3 \to \RR^3 $ that maps $I$ onto $J$ such that some similarity/dissimilarity $M(I\circ\phi, J)$ is optimized. Registration is an ill-posed problem. Therefore, the deformation $\phi$ requires regularization. Typical regularizations use constraints on the family of admissible transformations e.g. diffeomorphisms. Other alternatives are to enforced local constraints by using additional smoothing (enforcing scale to the transformation). The LOI and LOR framework provides building blocks for similarity measures, and do not impose regularisers forms. We use a very simple ones here.

\subsubsection*{Organisation and contributions.} 
The paper is organized as follows. First we review previous work in \cref{sec:relwork} and recall the Locally Orderless Imaging and Locally Orderless Registration frameworks in \cref{sec:loi}. Our main contribution, the extension of the LOI and LOR frameworks to first order information, is presented in \cref{sec:firstorder}. Registration objective functions are also discussed in this section. We illustrate the effects first order extensions for SSD and NCC similarities on the quality, and convergence of the registration in \cref{eq:exps}.  Finally, we summarise and discuss perspectives in  \cref{sec:conclusion}.
%
%
%


%% file: previous_work.tex
\section{Related Work}
\label{sec:relwork}
LOI was originally proposed by Koenderink and van Dorn \cite{koenderink1999structure} and describes the three inherent scales of images: spatial scale, intensity scale, and integration scale. This notion of images was used to describe image similarity in a variational framework \cite{hermosillo2002variational} and formalized into a generalized framework for image registration and the image similarity measures as LOR in \cite{darknersporring2012pami}. Some of the groundwork for LOR as well as the properties of the density estimators used for images in image registration where investigated in \cite{darknersporring2011ipmi}, revealing a 'scale imbalance' in the partial volume density estimator. The idea of marginalizing over more complex geometries than $\RR^n$ was proposed in \cite{jensen2015locally}. 

The idea of using higher order information for estimation of similarity between images is not new and normalized gradient fields NGF \cite{haber2006intensity} were one of the first. In \cite{sommer2013higher} an extension to the LDDMM using higher order information was presented.

There are few recent implementations of registration algorithms with NGF. The most noticeable uses NGF and a Gauss-Newton optimization scheme with locally rigid constraints \cite{konig2014fast}. This work was further evaluated on pelvis CT/CBCT images \cite{konig2016deformable}. A recent first-order information approach adds another metric based on gradients to the registration cost function with NGF \cite{theljani2019augmented}. This metric is defined as the sum of three gradients norms, i.e. the transformed moving image, the fixed image, and the difference between moving and fixed while offering a small increase in registration accuracy. 

%% file: background.tex
\section{Background on Locally Orderless Image Information}
\label{sec:loi}
\subsection{Notations}
$\Omega \subset \RR^3$ is the spatial domain of the images we use in the sequel. A scalar image is a function $f:\Omega\to \RR$. We assume that images can be extended out of $\Omega$ to $\RR^3$  -- typically by 0 -- as it is necessary for convolution. Convolution of two images $I,J:\RR^3$ is defined by $I*J(\bsx) = \int_{\RR^3}I(\bsy)J(\bsx-\bsy)\,d\bsy$. This actually extends to the case where one of the images is vector-valued directly.  
$G_\sigma$ denotes a 3D isotropic Gaussian of standard deviation $\sigma$.

\subsection{Lebesgue Integration and Histograms}

Consider a function integrable $I:\RR^n\to\RR$. Its integral $\int I\,d\mu$ with respect to the Lebesgue measure $\mu$ of $\RR^n$, denoted in the sequel as  $\int_{\RR^n} I(\bsx)\,d\bsx$,  can be computed as the limit over all subdivisions $0\leq i_0< \dots<i_N$ ,
$
\sum_{n=0}^{N-1} i_n \mu(I^{-1}([i_n,i_{n+1}]). 
$  
At the limit, when $i_{n+1}-i_n\to 0$, this can be rewritten as $\int_{\RR} i h_{I}(i)\, di$ where $h_{I}(i)$ is the length of isophote $I^{-1}(i)$. The function $i\mapsto h_{I}(i)$ is a generalized histogram of the values of $I$. Many standard integrals can be rewritten using this form. For instance 
$\int_{\RR^n} I(x)^2\,dx  = \int_{\RR} i^2 h_I(i)\,di$. This generalizes to joint histograms: given two images $I,J:\RR^n\to\RR$, $(I,J):\RR^n\to\RR^2$, $\bsx\mapsto (I(\bsx),J(\bsx))$ and its integral can be written as $\int_{\RR^2}(i,j)h_{I,J}\,di\,dj$ where $h_{I,J}$ is the joint histogram of $I$ and $J$. Classical similarities can be rewritten using histograms, for instance, Sum of Square Differences (SSD):  $\int_{\RR^n}(I(x)-J(x))^2\,d\bsx = \int_{\RR^2}(i-j)^2h_{I,J}(i,j)\,didj$.  Normalised Cross-Correlation, (Normalised) Mutual information, etc. can be written in terms of image histograms and their normalisations. 

\subsection{LOI and LOR framework}
LOI is a way to map images into local histograms, with three inherent scales: the spatial or image scale, the intensity scale, and integration scale. The image or spatial 
scale $\sigma$ is used to smooth input images $I$ and obtain $I_\sigma = I*G_\sigma$. A localised histogram over the values of $I_\sigma$ is computed as
\begin{equation}
     h_{I,\sigma\beta\alpha}(i|\bsx) :=  \int_\Omega P_\beta(I_\sigma(\bsy) - i) \, W_\alpha(\bsy - \bsx)\,d\bsy
     \label{eq:loi}
\end{equation}
where $P_\beta$ is a Parzen window of scale $\beta$, which provides the \emph{intensity} scale and $W_\alpha(x)$ is an integration window which provides the \emph{integration}  scale $\alpha$.
The histogram $h_{I,\sigma\beta\alpha}(\cdot|\bsx)$ is defined over $\RR$ or at least over an interval $\Lambda$ containing  the range of values of $I_\sigma$. 
Normalising it, we obtain the image density 
\begin{equation}
	\label{eq:density}
	p_{I,\sigma\beta\alpha}(i|\bsx) = \frac{h_{I,\sigma\beta\alpha}(i|\bsx)}{\int_\Lambda h_{I,\sigma\beta\alpha}(j|\bsx)\,dj}.
\end{equation}
By letting the integration scale $\alpha\to\infty$, we obtain \emph{global} histograms an densities	$h_{I,\sigma\beta}(i) :=  \int_\Omega P_\beta(I_\sigma(\bsx) - i)\,dx$ and  $p_{I,\sigma\beta}(i)$. This will be the case in this paper.
This construction extends to the definition of joint histograms and densities, at the heart of Locally Orderless Registration by
\begin{align}
	h_{I,J,\sigma\beta\alpha}(i,j|\bsx) &:= \int_\Omega  P_\beta(I_\sigma(\bsy) - i) P_\beta(J_\sigma(\bsy) - j) \, W_\alpha(\bsy - \bsx)\,d\bsy\label{eq:hist2}\\
	p_{I,J,\sigma\beta\alpha}(i,j|\bsx) &= \frac{h_{I,J,\sigma\beta\alpha}(i,j|\bsx)}{\int_\Lambda h_{I,J,\sigma\beta\alpha}(u,v|\bsx)\,du\,dv}\label{eq:joinddensity}
\end{align}
and similar formulas  in the global case. Single histograms and densities can also be obtained from them by marginalisation. LOR Image similarities are defined through single and joint density estimates \cref{eq:density} and \cref{eq:joinddensity}. 
Similarity measures are defined as 
\begin{align}
	M_L(I,J) &= \int_\Omega\int_{\Lambda^2}f(i,j,p_{I,J,\sigma\beta\alpha}(i,j|\bsx))\,di\,dj\,d\bsx,\\
	M_G(I,J) &= \int_{\Lambda^2}f(i,j,p_{I,J,\sigma\beta}(i,j))\,di\,dj
\end{align}
with $M_L$ built from localised densities and $M_G$ from global ones. 
Among them, $p-$linear ones are characterized by $f(i,j,p) = g(i,j)p$, while nonlinear ones take more complex forms. We already mentioned in the previous section how SSD can be simply written using joint histograms. By normalising it, it can be written via densities \eqref{eq:joinddensity}. Another classical similarity, normalised cross-correlation (NCC), can also easily be written in term of histograms and densities. 
$$
NCC(I,J) = \frac{\iprod{I-\bar{I}}{J-\bar{J}}}{\|I-\bar{I}\|\|J-\bar{J}\|}
$$ 
where $\bar{I}$ and $\bar{J}$ are the average values of $I$ and $J$ on $\Omega$ and the inner product and norms are $L^2$ ones. The inner product $\iprod{I}{J}$ is $\int_{\RR^2}ij h_{I,J}(i,j)\,didj$. Replacing $h_{I,J}$ by $h_{I,J;\sigma\beta\alpha}$ provides its LOI counterpart expression. The average $\bar{I}$ is $\int_{\RR}ih_I(i)\,di/(\int_{\RR}h(i)\,di)$. Again, we replace  $h_I$ by $h_{I;\sigma\beta\alpha}$ to obtain its LOI counterpart expression. 

To use it in registration, the setting is typically the following. One chooses a hold on domain $D\subset\RR^3$ large enough, with $\Omega\subset D$ 
and mappings $\phi:\RR^3\to\RR^3$, with $\phi\equiv \Id_3$ out of $D$, where $\Id_3$ is the identity transform. 
Here, we assume $D=\Omega$. These transformations are usually of class $\Cc^k$, $k\geq 1$, often more.
They are often, but not always, constrained to be diffeomorphic. A goodness of fit functional is obtained by evaluation the (dis)similarity  $\phi\mapsto M(I\circ\phi, J)$.

%
%
%
%

%% file: method.tex
\section{Extension of LOI and LOR to Higher Information}
\label{sec:firstorder}
In this section, we introduce a straightforward way to extend the LOI to incorporate higher order image information in histogram and density formulations. We focus on first order, as higher order may be limited in practice because of the complexity and resulting memory footprint.

\subsection{First Order Locally Orderless Registration (FLOR)}
In this paper we probe and use first order differential information of an image $I:\RR^3\to\RR$ (with effective spatial domain $\Omega$). It is obtained by lifting it to image $\bsI_{\sigma}:\RR^3\times\SS^2\to\RR$
which encodes gradient responses at different directions in a straightforward way. The differential $d_{\bsx}G_\sigma:\RR^3\to \RR$ is, for each $\bsx$ linear, it is enough to know it on $\SS^2\subset \RR^3$.
\begin{equation}
	\label{eq:lifting}
	\bsI_\sigma(\bsx,\bsv) = \left(\int_{\RR^3}I(\bsy)d_{(\bsy-\bsx)}G_\sigma\,d\bsy\right)\bsv = d_{\bsx}\left(I*G_\sigma \right)\bsv
\end{equation}
This can of course be rewritten as $\bsI_\sigma(\bsx,\bsv) = \nabla I_\sigma(\bsx)^T\bsv$. Note that $\bsI_\sigma(\bsx,-\bsv) =  -\bsI_\sigma(\bsx,\bsv)$ due to our lifting choice.
Using a higher order operator, such as , for instance, the Hessian of Gaussian $\nabla^2 G_\sigma$ would allow us to probe second order structure as a $\tilde{\bsI}_\sigma(\bsx,\bsv) = \Hess I_\sigma(\bsx)(\bsv, \bsv)$.

Once the lifting is performed, we can now define local histograms and densities. They are spatially localised, not directionally.
\begin{align}
	h_{\bsI;\sigma\beta\alpha}(i|\bsx) &= \int_{\RR^3\times\SS^2}P_\beta(\bsI_\sigma(\bsy,\bsv) - i)W_\alpha(\bsx-\bsy)\,d\bsv\,d\bsy\\
	p_{\bsI;\sigma\beta\alpha}(i|\bsx) &= \frac{h_{\bsI;\sigma\beta\alpha}(i|\bsx)}{\int_\Lambda h_{\bsI;\sigma\beta\alpha}(j|\bsx)\,dj}
\end{align}
where this time $\Lambda$ is an interval containing the range of $\bsI_\sigma$. As in the zeroth order case, global histograms and densities can be obtained by letting $\alpha\to\infty$.
Given two images $I,J:\RR^3\to\RR$, we can lift them to $\bsI_\sigma$ and $\bsJ_\sigma$ and define joint histograms and densities
\begin{align}
	h_{\bsI,\bsJ;\sigma\beta\alpha}(i,j|\bsx) &= \int_{\RR^3\times\SS^2}\!\!\!\!\!\!P_\beta(\bsI_\sigma(\bsy,\bsv) \!-\! i)P_\beta(\bsJ_\sigma(\bsy,\bsv) \!-\! i)W_\alpha(\bsx\!-\!\bsy)\,d\bsv\,d\bsy\\
	p_{\bsI,\bsJ;\sigma\beta\alpha}(i,j|\bsx) &= \frac{h_{\bsI;\bsJ\sigma\beta\alpha}(i,j|\bsx)}{\int_{\Lambda^2} h_{\bsI;\sigma\beta\alpha}(u, v|\bsx)\,du\,dv}
\end{align}
Here again, by letting $\alpha\to 0$, we obtain global histograms and densities.

\subsection{First Order Deformation Model}
Let $\phi:\RR^3\to\RR^3$ a deformation. By the chain rule, $\dirdert{}I_\sigma(\phi(\bsx+t\bsv)) = \nabla I_\sigma(\phi(\bsx))^T J_{\bsx}\phi(\bsv)$, with $J_{\bsx}\phi$ the Jacobian of $\phi$. This implies of course that $\phi$ acts on the first order information via its differential. Here comes the problem that, as we have limited the directional probing space to $\SS^2$, there is no guarantee that $J_{\bsx}\phi(\bsv)\in\SS^2$, let alone non zero. This is however the case if we restrict $\phi$ to be a diffeormorphism, and this is what we assume from now. From its very definition, the mapping of directions at $\bsx\in\Omega$ is given by
\begin{equation}
	\psi_{\bsx}:\SS^2\to\SS^2,\quad v \mapsto \frac{J_{\bsx}\phi(\bsv)}{|J_{\bsx}\phi(\bsv)|}, 
\end{equation}
This lead to define the action of $\phi$\footnote{Note that this is not stricto senso a group action here.} on the lifted image $\bsI_\sigma(\bsx,\bsv)$ as
\begin{equation}
\label{eq:scaling}
	\left(\phi.\bsI_\sigma\right)(\bsx,\bsv)= |J_{\bsx}\phi(\bsv)|\bsI_\sigma(\phi(\bsx),\psi_{\bsx}(\bsv)).
\end{equation}   
It clearly satisfies $\left(\phi.\bsI_\sigma\right)(\bsx,-\bsv) = -\left(\phi.\bsI\right)(\bsx,\bsv)$, thus respecting the structure of lifted images. 
Alternatively, one could consider another first order deformation model, where the Jacobian scaling factor is ignored, i.e. 
\begin{equation}
\label{eq:noscaling}
\left(\phi.\bsI_\sigma\right)(\bsx,\bsv) = \bsI_\sigma(\phi(\bsx),\psi_{\bsx}(\bsv)).
\end{equation}
This may apply to images of more categorical nature. This can be the case for two images showing similar anatomical structures, with same tissue density, but which cannot be registered by a (local) rigid motion.

By using either the local or global histograms and densities,  
higher order similarities $\bsM(\bsI,\bsJ)$ are obtained exactly the same way as discussed in the previous section. 
Finally one may combine zeroth and first order to get new similarity measures, and use them in a registration framework via
\begin{equation}
	\phi\mapsto M(I\circ\phi,J) + \lambda \bsM(\phi.\bsI_\sigma,\bsJ_\sigma).
\end{equation}

\subsection{Registration Objectives and Deformations.}
The similarities used in this paper are 1) SSD for zeroth and first order information, and 2) NCC for zeroth and first order information.
Free-form B-spline deformation models \cite{rueckert1999nonrigid} are used, with simple control point grid motion limitation as regularisation. We also use simple translation deformations on some experiments.

\subsection{Implementation}
The implementation has been made in PyTorch 1.7.1 and the basis consists of a Cubic B-spline from which both the image interpolation and deformation field can be estimated. Analytical Jacobians of both image and deformation has been implemented which allow us to use the backpropagation of PyTorch and optimizer for finding the solution. The action of the Jacobian on the directional derivative have 2 implementations, given by \Cref{eq:scaling} and \Cref{eq:noscaling}. The implementations ensures that all scales are consistent and to change image scale we simply blur the images prior to registration with the desired kernel. 
Objectives are optimised using PyTorch Adam implementation. The code runs both on CPU and GPU; a full 3D registration takes 2-3 minutes on a laptop and around 1 minute on an RTX3090.  

%% file: results.tex
\section{Experiments \& Results}
\label{eq:exps}
\begin{figure}
\centering
\begin{subfigure}{.35\linewidth}
    \centering
I    \includegraphics[width=0.9\linewidth]{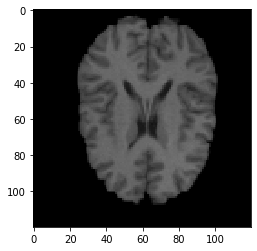}
    \caption{Target}
\end{subfigure}
\begin{subfigure}{.35\linewidth}
    \centering
    \includegraphics[width=0.9\linewidth]{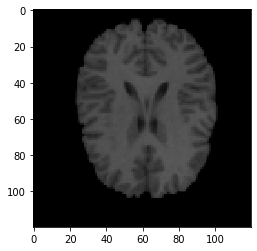}
    \caption{Source}
\end{subfigure}
    \caption{A Slice of the target and source image used for our experiments}
    \label{fig:subjects}
\end{figure}

We conduct 2 main experiments. First we investigate the properties of the $1^{st}$-order information compared to the $0^{th}$-order using translation only. Secondly we show that we can perform 3D non-rigid registration with convincing results.
We have used two 3D T1 weighted magnetic resonance images (MRI) from two separate individuals for our proof of principle. The images are shown in  \Cref{fig:subjects}.
\subsection{The similarity properties}
To illustrate the effects of including higher-order ($1^{st}$-order) information in the similarity-measure,
we map the $0^{th}$-order and $1^{st}$-order information  as a function of translation in 2D $(x,y)$-plane. 

\begin{figure}
\begin{subfigure}{.5\linewidth}
    \centering
    \includegraphics[ width=0.9\linewidth]{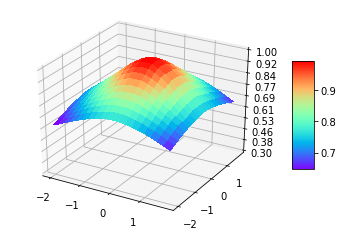}
    \caption{NCC of 0th}
\end{subfigure}
\begin{subfigure}{.5\linewidth}
    \centering
    \includegraphics[ width=0.9\linewidth]{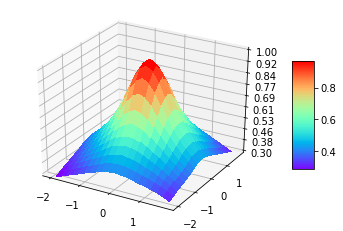}
    \caption{NCC of 1st}
\end{subfigure}
\begin{subfigure}{.5\linewidth}
    \centering
    \includegraphics[ width=0.9\linewidth]{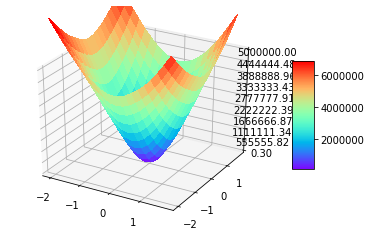}
    \caption{SSD of 0th}
\end{subfigure}
\begin{subfigure}{.5\linewidth}
    \centering
    \includegraphics[ width=0.9\linewidth]{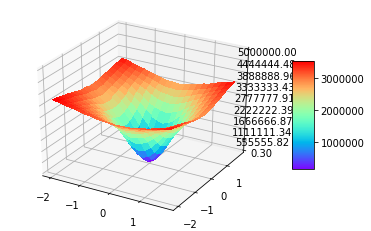}
    \caption{SSD of 1st}
\end{subfigure}
    \caption{The self similarity of the 0th and 1st order image information for NCC and SSD respectively under translation in 2D around the identity.}
     \label{fig:optim}
\end{figure}

\begin{figure}
\begin{subfigure}{.5\linewidth}
    \centering
    \includegraphics[ width=0.9\linewidth]{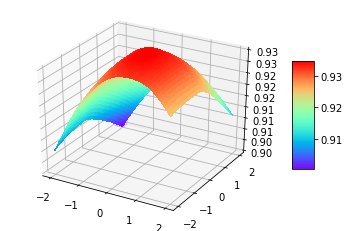}
    \caption{NCC of 0th}
\end{subfigure}
\begin{subfigure}{.5\linewidth}
    \centering
    \includegraphics[ width=0.9\linewidth]{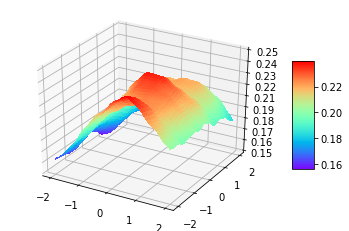}
    \caption{NCC of 1st}
\end{subfigure}
\begin{subfigure}{.5\linewidth}
    \centering
    \includegraphics[ width=0.9\linewidth]{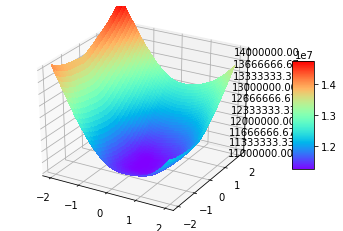}
    \caption{SSD of 0th}
\end{subfigure}
\begin{subfigure}{.5\linewidth}
    \centering
    \includegraphics[ width=0.9\linewidth]{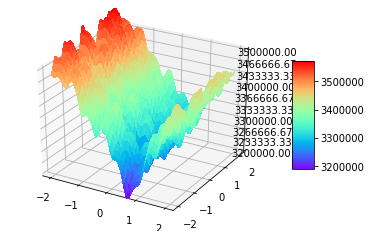}
    \caption{SSD of 1st}
\end{subfigure}
    \caption{The similarity of the $0^{th}$- and $1^{st}$-order image information for NCC and SSD respectively under translation in 2D around the identity for the source and target image. Clearly,
     multiple local minima exist in the $1^{st}$-order information, in contrast to $0^{th}$-order that only has one.}
    \label{fig:test_different}
\end{figure}

Our first experiment shows how the information from the images using SSD and NCC respectively appears in the simple case where the deformation $\phi$ is a pure translation in 2D, for an MRI compared with itself. As \Cref{fig:optim} illustrates, the similarity in the $1^{st}$-order information has a significantly steeper slope close to the optimum, compared to that of the $0^{th}$-order information for both SSD and NCC. This indicates that including $1^{st}$-order information may improve registration close to the optimum. However, when comparing 2 different images in \Cref{fig:test_different} we observe that multiple minima exist with $1^{st}$-order only, and that $0^{th}$-order has a better and a wider basin of attraction. Furthermore NCC seems to be more suitable compared to SSD. Therefore a combination of  $0^{th}$-order, $1^{st}$-order information and NCC seems more appropriate for image registration applications.

\subsection{Non-rigid registration}
\begin{figure}
\begin{subfigure}{.5\linewidth}
    \centering
    \includegraphics[ width=0.9\linewidth]{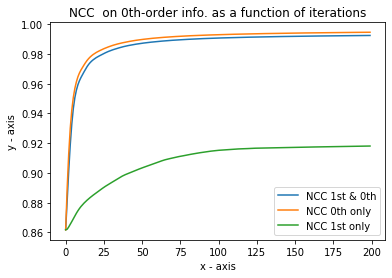}
    \caption{NCC of 0th}
\end{subfigure}
\begin{subfigure}{.5\linewidth}
    \centering
    \includegraphics[ width=0.9\linewidth]{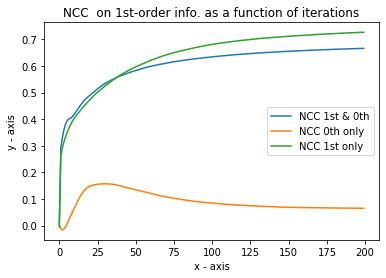}
    \caption{NCC of 1st}
\end{subfigure}
    \caption{Convergence plot of NCC for $0^{th}$- and $1^{st}$-order similarities, separately, as function of iterations. The experiment was performed optimizing only for $0^{th}$-order information, only $1^{st}$-order information and both with weight between $0^{th}$- and $1^{st}$-order terms as the ratio of the number of evaluation-orientations ({\small $\frac{1}{26}$}). Note how the 1st-order appears to also maximize the $0^{th}$-order and how $1^{st}$-order fails to maximize the gradient information.}
    \label{fig:conv}
\end{figure}

We perform 3 non-rigid registrations of the source and the target using only $1^{st}$-order information and using only $0^{th}$-order information and a combination of both respectively. We used a free-form deformation cubic B-spline~\cite{rueckert1999nonrigid} with 5 voxel spacing between the knots and evaluation points for every second voxel. We discretized the  $1^{st}$-order information with 26 normalized directions, pointing to each neighbouring voxel in a $3\times3\times 3$ local grid. We weighted $0^{th}$-order and $1^{st}$-order terms by the ratio between the number evaluation of the $0^{th}$-order and the $1^{st}$-order ({\small $\frac1{26}$}). As can be seen from the convergence plots (\Cref{fig:conv}), this ratio will align both gradient information and intensity information, in contrast to  optimizing only the $0^{th}$-order or the $1^{st}$-order information. 

\begin{figure}

\begin{subfigure}{.32\linewidth}
    \centering
    \includegraphics[ width=0.99\linewidth]{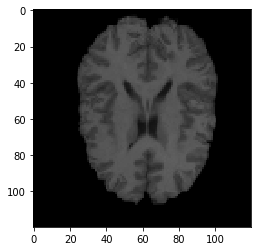}
    \caption{Deformed source.}
\end{subfigure}
\begin{subfigure}{.32\linewidth}
    \centering
    \includegraphics[ width=0.99\linewidth]{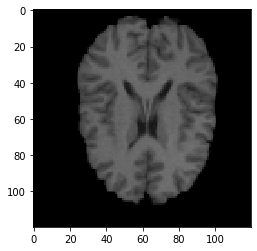}
    \caption{Target}
\end{subfigure}
\begin{subfigure}{.32\linewidth}
    \centering
    \includegraphics[ width=0.99\linewidth]{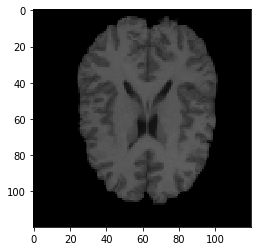}
    \caption{Deformed source.}
\end{subfigure}
\centering
\begin{subfigure}{.41\linewidth}
    \centering
    \includegraphics[ width=0.99\linewidth]{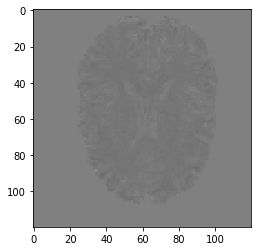}
    \caption{Difference.}
\end{subfigure}
\begin{subfigure}{.41\linewidth}
    \centering
    \includegraphics[ width=0.99\linewidth]{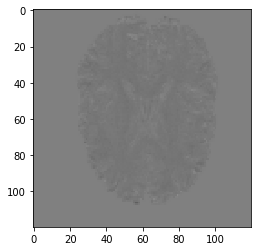}
    \caption{Difference.}
\end{subfigure}
\\
\centering
\begin{subfigure}{.41\linewidth}
    \centering
    \includegraphics[ width=0.99\linewidth]{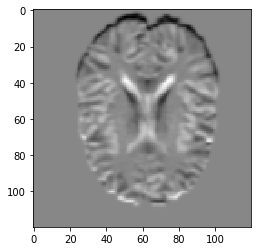}
    \caption{Gradient information for 1 direction, no scaling}
\end{subfigure}
\begin{subfigure}{.41\linewidth}
    \centering
    \includegraphics[ width=0.99\linewidth]{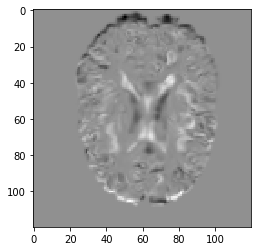}
    \caption{Gradient information for 1 direction, scaling}
\end{subfigure}
    \caption{(a) and (c) are the result of the registration to (b), where (a) has been registered with \Cref{eq:scaling} and (c) with \Cref{eq:noscaling}. (d) and (e) are the difference images of (a) and (b),  and  (c) and (b) respectively. (f) and (g) are examples of the 1st order information in (c) matched to (b) in the frame of (c)}
    \label{fig:nonrigid_reg}
\end{figure}

The final registration results are shown in \Cref{fig:nonrigid_reg}. We have used both formulations from \cref{eq:scaling}, 
with Jacobian normalisation, and \cref{eq:noscaling},  without Jacobian normalisation. As \Cref{fig:nonrigid_reg} shows, 
the results are quite convincing, and the difference between the two is very small. However, in this MRI registration case, 
the version not including scaling seems to be more suitable compared to SSD. 

%% file: conclusion.tex
\subsection{Discussion and Limitations}
\label{sec:conclusion}
\begin{figure}[htb!]
    \centering
    \includegraphics[ width=0.25\linewidth]{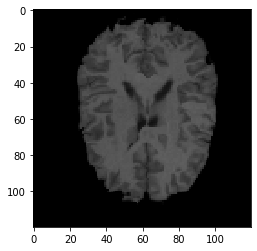}
    \caption{An example of the deformation from pure 1st-order information.}
    \label{fig:1st_reg}
\end{figure}

The experiments presented in the previous section illustrate the gain obtained by including  $1^{st}$-order information, but also some limitations. First, it is clear from our experiments that $1^{st}$-order information itself is not sufficient to make a proper registration in our setup. \Cref{fig:1st_reg} shows it clearly, where the registration of the image is inferior and suffers from undesirable deformations. Another point is that  experiment here only serve as a proof of principle, and future work will include a thorough comparison over multiple large data sets similar to \cite{klein2009evaluation}. Furthermore, we will extend the work to use diffeomorphisms like previous methods, such as: Symmetric Normalization \cite{avants2008symmetric} or the Collocation for Diffeormorphic Deformations \cite{darkner2018collocation}. Finally, we will include an evaluation of information theory metrics and the exploration of all the scales presented in the formulation.

\section{Conclusion}

We introduced a framework for including higher-order information into image similarity and illustrated the application in image registration. We have shown that the method is able to match both $0^{th}-$ and $1^{st}-order$ information using SSD and NCC. The framework allows us to use all admissible measures from the LOR framework. We have shown that the framework is able to deliver high-quality non-rigid registration and that it has the potential to improve the accuracy of image registration in general.

%% file: acknowledgements.tex
\section{Acknowledgements}

\begin{wrapfigure}{R}{.13\linewidth}
\def\svgwidth{\linewidth}
\vspace*{-2em}
\centering
\includegraphics[width=.99\linewidth]{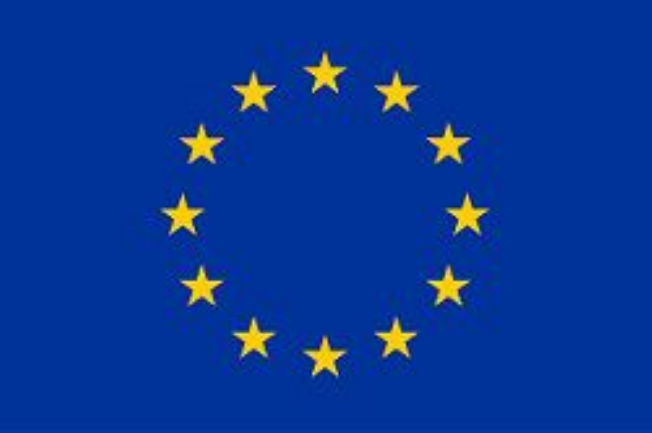}
\end{wrapfigure}  
This project has received funding from the European Union's Horizon 2020 research and innovation programme under the Marie Sk\l{}odowska-Curie grant agreement No. 764644.
This paper only contains the author's views and the Research Executive Agency and the Commission are not responsible for any use that may be made of the information it 

%% file: main.bbl
\begin{thebibliography}{10}
\providecommand{\url}[1]{\texttt{#1}}
\providecommand{\urlprefix}{URL }
\providecommand{\doi}[1]{https://doi.org/#1}

\bibitem{avants2008symmetric}
Avants, B.B., Epstein, C.L., Grossman, M., Gee, J.C.: Symmetric diffeomorphic
  image registration with cross-correlation: evaluating automated labeling of
  elderly and neurodegenerative brain. Medical image analysis  \textbf{12}(1),
  26--41 (2008)

\bibitem{darkner2018collocation}
Darkner, S., Pai, A., Liptrot, M., Sporring, J.: Collocation for diffeomorphic
  deformations in medical image registration. I E E E Transactions on Pattern
  Analysis and Machine Intelligence  \textbf{40}(7),  1570--1583 (2018).
  \doi{10.1109/TPAMI.2017.2730205}

\bibitem{darknersporring2011ipmi}
Darkner, S., Sporring, J.: Generalized partial volume: An inferior density
  estimator to parzen windows for normalized mutual information. In: IPMI.
  LNCS, vol.~6801, pp. 436--447. Springer (2011)

\bibitem{darkner2013locally}
Darkner, S., Sporring, J.: Locally orderless registration. IEEE transactions on
  pattern analysis and machine intelligence  \textbf{35}(6),  1437--1450 (2013)

\bibitem{darknersporring2012pami}
Darkner, S., Sporring, J.: {Locally Orderless Registration}. IEEE Transactions
  on Pattern Analysis and Machine Intelligence  \textbf{35}(6),  1437--1450
  (2013)

\bibitem{haber2006intensity}
Haber, E., Modersitzki, J.: Intensity gradient based registration and fusion of
  multi-modal images. In: International Conference on Medical Image Computing
  and Computer-Assisted Intervention. pp. 726--733. Springer (2006)

\bibitem{hermosillo2002variational}
Hermosillo, G., Chefd'Hotel, C., Faugeras, O.: {Variational methods for
  multimodal image matching}. International Journal of Computer Vision
  \textbf{50}(3),  329--343 (2002)

\bibitem{janssen_etal:2018}
Janssen, M., Janssen, A., andJ. Oliv{\'a}n~Bescos, E.B., Duits, R.: {Design and
  Processing of 3D invertible Orientation Scores of 3D Images}. Journal of
  Mathematical Imaging and Vision  \textbf{60},  1427--1458 (2018)

\bibitem{jensen2015locally}
Jensen, H.G., Lauze, F., Nielsen, M., Darkner, S.: Locally orderless
  registration for diffusion weighted images. In: International Conference on
  Medical Image Computing and Computer-Assisted Intervention. pp. 305--312.
  Springer (2015)

\bibitem{klein2009evaluation}
Klein, A., Andersson, J., Ardekani, B.A., Ashburner, J., Avants, B., Chiang,
  M.C., Christensen, G.E., Collins, D.L., Gee, J., Hellier, P., et~al.:
  Evaluation of 14 nonlinear deformation algorithms applied to human brain mri
  registration. Neuroimage  \textbf{46}(3),  786--802 (2009)

\bibitem{koenderink1999structure}
Koenderink, J.J., Van~Doorn, A.J.: The structure of locally orderless images.
  International Journal of Computer Vision  \textbf{31}(2),  159--168 (1999)

\bibitem{konig2016deformable}
K{\"o}nig, L., Derksen, A., Papenberg, N., Haas, B.: Deformable image
  registration for adaptive radiotherapy with guaranteed local rigidity
  constraints. Radiation Oncology  \textbf{11}(1), ~1--9 (2016)

\bibitem{konig2014fast}
K{\"o}nig, L., R{\"u}haak, J.: A fast and accurate parallel algorithm for
  non-linear image registration using normalized gradient fields. In: 2014 IEEE
  11th international symposium on biomedical imaging (ISBI). pp. 580--583. IEEE
  (2014)

\bibitem{rueckert1999nonrigid}
Rueckert, D., Sonoda, L.I., Hayes, C., Hill, D.L., Leach, M.O., Hawkes, D.J.:
  Nonrigid registration using free-form deformations: application to breast mr
  images. IEEE transactions on medical imaging  \textbf{18}(8),  712--721
  (1999)

\bibitem{sommer2013higher}
Sommer, S., Nielsen, M., Darkner, S., Pennec, X.: Higher-order momentum
  distributions and locally affine lddmm registration. SIAM Journal on Imaging
  Sciences  \textbf{6}(1),  341--367 (2013)

\bibitem{theljani2019augmented}
Theljani, A., Chen, K.: An augmented lagrangian method for solving a new
  variational model based on gradients similarity measures and high order
  regularization for multimodality registration. Inverse Problems and Imaging
  (2019)

\end{thebibliography}
